\documentclass{article}

\usepackage{tabularx}
\usepackage{natbib}
\usepackage{tablefootnote}
\usepackage[table]{xcolor}
\usepackage[normalem]{ulem}

\definecolor{lightgray}{gray}{0.9}
\usepackage[final]{neurips_2023}

\usepackage[utf8]{inputenc} % allow utf-8 input
\usepackage[T1]{fontenc}    % use 8-bit T1 fonts
\usepackage{hyperref}       % hyperlinks
\usepackage{url}            % simple URL typesetting
\usepackage{booktabs}       % professional-quality tables
\usepackage{amsfonts}       % blackboard math symbols
\usepackage{nicefrac}       % compact symbols for 1/2, etc.
\usepackage{microtype}      % microtypography
\usepackage{xcolor}         % colors
\usepackage{booktabs} 
\usepackage{microtype}
\usepackage{amsmath}
\usepackage{comment}
\usepackage{enumitem}
\usepackage{graphicx}
\usepackage[export]{adjustbox}[2011/08/13]
\usepackage{wrapfig}
\usepackage{caption}

\newcommand{\polym}{\texttt{Poly}-$\mu$}
\newcommand{\polysm}{\texttt{MHR}-$\mu$}
\newcommand{\random}{\texttt{Random}-$\mu$}

\newcommand{\AS}{\texttt{AdapterSoup}}
\newcommand{\poly}{\texttt{Poly}}
\newcommand{\tfew}{\texttt{T-Few}}
\newcommand{\shared}{\texttt{LoRA}}
\newcommand{\polys}{\texttt{MHR}}

\usepackage{multirow}
\usepackage{arydshln}
\makeatletter
\def\adl@drawiv#1#2#3{%
        \hskip.5\tabcolsep
        \xleaders#3{#2.5\@tempdimb #1{1}#2.5\@tempdimb}%
                #2\z@ plus1fil minus1fil\relax
        \hskip.5\tabcolsep}
\newcommand{\cdashlinelr}[1]{%
  \noalign{\vskip\aboverulesep
           \global\let\@dashdrawstore\adl@draw
           \global\let\adl@draw\adl@drawiv}
  \cdashline{#1}
  \noalign{\global\let\adl@draw\@dashdrawstore
           \vskip\belowrulesep}}
\makeatother
\title{Multi-Head Adapter Routing \\for Cross-Task Generalization}
\usepackage{amsmath,amsfonts,bm}

\def\eqref#1{equation~\ref{#1}}
\def\1{\bm{1}}

\def\vh{{\bm{h}}}

\def\vl{{\bm{l}}}

\def\mA{{\bm{A}}}
\def\mB{{\bm{B}}}

\def\mW{{\bm{W}}}

\def\mZ{{\bm{Z}}}

\DeclareMathAlphabet{\mathsfit}{\encodingdefault}{\sfdefault}{m}{sl}
\SetMathAlphabet{\mathsfit}{bold}{\encodingdefault}{\sfdefault}{bx}{n}
\newcommand{\tens}[1]{\bm{\mathsfit{#1}}}

\def\tZ{{\tens{Z}}}

\def\emZ{{Z}}

\newcommand{\etens}[1]{\mathsfit{#1}}

\def\etZ{{\etens{Z}}}

\newcommand\crule[1][black]{\textcolor{#1}{\rule{2mm}{2mm}}}
\definecolor{myred}{HTML}{ff0000}
\definecolor{mygreen}{HTML}{008000}
\definecolor{myviolet}{HTML}{ee82ee}
\definecolor{myyellow}{HTML}{ffff00}
\definecolor{myindigo}{HTML}{4b0082}
\definecolor{myorange}{HTML}{ffa500}
\definecolor{myblue}{HTML}{0000ff}

\author{
Lucas Caccia$^{\crule[myred]\crule[mygreen]\crule[myblue]}$
\qquad Edoardo Ponti$^{\crule[myorange]}$
\qquad Zhan Su$^{\crule[myviolet]}$
\qquad Matheus Pereira$^{\crule[myred]}$ \vspace{.3cm} \\
\qquad \textbf{Nicolas Le Roux}$^{\crule[myred]\crule[mygreen]\crule[myblue]\crule[myindigo]}$
\qquad \textbf{Alessandro Sordoni}$^{\crule[myred]\crule[myblue]\crule[myindigo]}$
\vspace{.4cm} \\
$^{\crule[myred]}$Microsoft Research, $^{\crule[mygreen]}$McGill University, $^{\crule[myblue]}$MILA, \vspace{0.05cm} \\
$^{\crule[myorange]}$University of Edinburgh, $^{\crule[myindigo]}$Universit\'e de Montr\'eal, $^{\crule[myviolet]}$University of Copenhagen \vspace{0.1cm} \\
\texttt{lucas.page-caccia@mail.mcgill.ca,alsordon@microsoft.com}
}

\begin{document}

\maketitle

\begin{abstract}
Parameter-efficient fine-tuning (PEFT) for cross-task generalization consists in pre-training adapters on a multi-task training set before few-shot adaptation to test tasks. Polytropon~\citep{ponti2022combining} (\poly{}) jointly learns an inventory of adapters and a \emph{routing} function that selects a (variable-size) subset of adapters for each task during both pre-training and few-shot adaptation. In this paper, we investigate the role that adapter routing plays in its success and design new variants based on our findings.
First, we build on the intuition that finer-grained routing provides more expressivity. Hence,
we propose \polys{} (Multi-Head Routing), which combines \textit{blocks} of parameters from different adapters and outperforms \poly{} under a comparable parameter budget; by only fine-tuning the routing function and not the adapters (\polys{}-$z$), we achieve competitive performance with extreme parameter efficiency. Second, we find that \poly{}/\polys{} performance is a result of better multi-task optimization, rather than modular inductive biases that facilitate adapter recombination and local adaptation, as previously hypothesized. In fact, we find that \polys{} exhibits high gradient alignment between training tasks. We find that routing is most beneficial during multi-task pre-training rather than during few-shot adaptation and propose \polysm{}, which discards routing and fine-tunes the average of the pre-trained adapters on each downstream tasks. This establishes \polysm{} as an effective method for single-adapter fine-tuning. We also show that \polysm{} can be used as an effective zero-shot transfer method by training the average of the pre-trained adapters for a few additional steps on the multi-task training set: this yields gains up to 3\% on absolute accuracy w.r.t. the baselines. Code is available at \url{https://github.com/microsoft/mttl}.
\end{abstract}

\begin{figure}[t]
\centering
\includegraphics[width=0.99\linewidth]{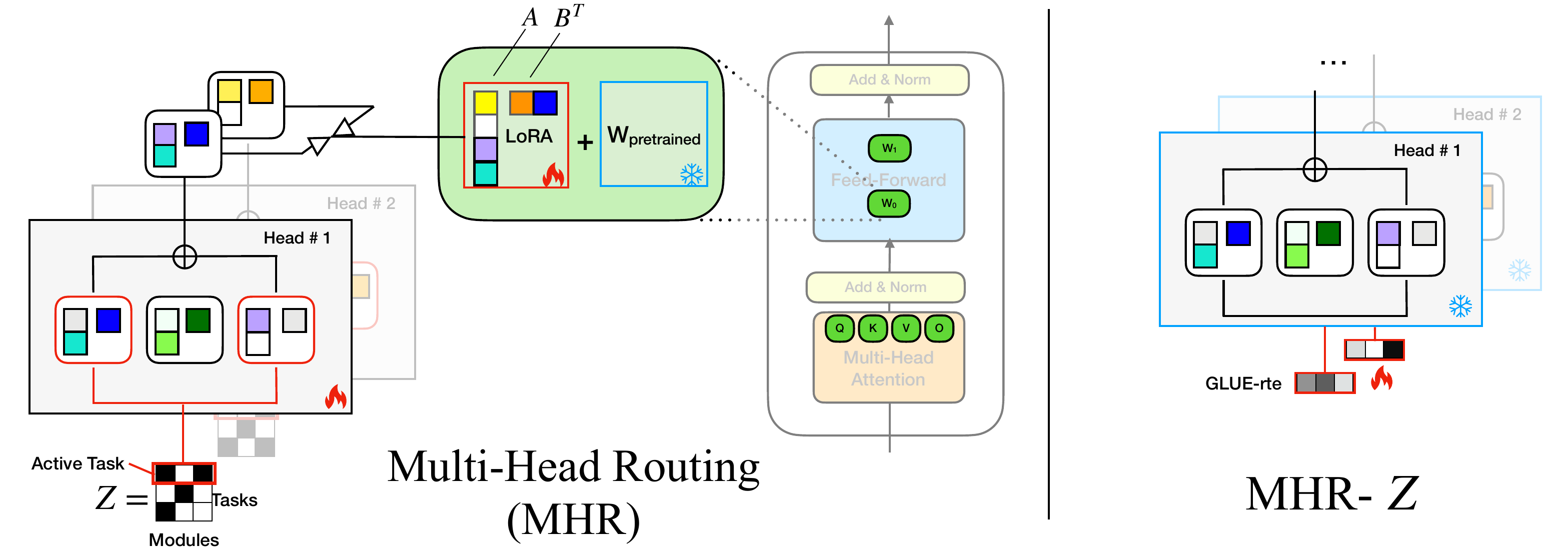}
\caption{\textit{Left:} A LoRA adapter with weight $A B^\top$ is trained on top of a frozen, pre-trained linear layer $W$. Our method \polys{} partitions the $A, B$ parameter indexes into $h$ blocks (or \textit{heads}). For each block, a separate routing function selects the active modules for the current task among $m$ copies with different parameter values, and combines them via averaging to form a task-specific head. The heads are then concatenated to form the LoRA adapter. Using multiple heads allows for more fine-grained mixing of task parameters with a negligible increase in overall parameter count. \textit{Right:} During few-shot adaptation, one can fine-tune only the multi-head routing parameters (\polys{}-$z$), keeping the modules frozen, resulting in highly parameter-efficient adaptation.}
\label{fig:first_fig}
\end{figure}

\section{Introduction}
The ability to train effective models with a relatively small number of training data is of paramount importance due to the paucity of annotated examples for most tasks.
One effective few-shot learning approach is to leverage large models pre-trained on a vast amount of unlabelled data and fine-tune them on the few examples available for each downstream task.
To reduce the memory cost of duplicating the entire array of parameters for each downstream task, recent approaches resort to parameter-efficient fine-tuning (PEFT) methods, such as LoRA~\citep{lora}, SFT \citep{ansell2021composable}, or (IA)$^3$~\citep{tfew}. These only fine-tune adapters while leaving the pre-trained model `frozen'.

Nevertheless, it remains unclear how to best exploit a set of \emph{training} tasks to better generalize to a set of unseen \emph{test} tasks in a sample-efficient fashion, based on just a few examples. One straightforward solution is to perform multi-task pre-training,~i.e.\ first train the large model on the union of the examples from the training tasks, then fine-tune the obtained model to the test task~\citep{tfew,xfit}. However, this solution does not take into account that test tasks may require solving different combinations of sub-problems compared to training tasks~\citep{vu-etal-2020-exploring}, thus failing to achieve compositional generalization \citep{rosenbaum2019routing,ponti2021inductive}. Moreover, specializing the model towards different tasks during training may result in negative transfer, due to their corresponding gradients being misaligned \citep{wang2021gradient}. 

Several PEFT approaches have been proposed to enable better cross-task generalization by training adapters (or soft prompts) on each task independently~\citep{pfeiffer2020adapterfusion, vu2021spot, asai2022attempt,chronopoulou2023adaptersoup}. Given a new test task, parameters from similar training tasks are aggregated, which enables transfer. While solely having task-specific parameters is an effective strategy to mitigate interference across training tasks, it also inhibits any positive transfer within the same task pool.
Polytropon (\texttt{Poly}) was recently proposed by \citet{ponti2022combining} to address these issues: the model assumes that task-specific adapters are learned combinations of a reusable inventory of basis adapters or \emph{modules}. In practice, each module is implemented as a LoRA~\citep{lora} adapter, which modifies a large pre-trained model, such as T5~\citep{t5}.
During both multi-task pre-training and few-shot adaptation, \poly{} learns both the inventory of adapters and a (continuously relaxed) binary task--module routing matrix, which determines which module is active for each task. While \poly{} shows promising results, several questions remain unanswered: 1) Does the expressivity of the routing function matter? 2) Why do routing-based PEFT methods yield superior performance? 3) Is routing useful during both multi-task pre-training and few-shot adaptation?

To answer the first question, we propose a new routing function, \polys{}, that mixes adapters at a more granular level. Differently from \poly{}, where routing decisions are made for each adapter as a whole, in \polys{} we linearly combine blocks of the adapter dimensions (i.e. heads), each with different combination coefficients.  
% This comes at negligible extra costs in terms of parameter count during multi-task training. During inference, the resulting composition shares the same structure and parameter count as the PEFT method used for each individual skill, and therefore comes at no additional compute or memory cost.
We evaluate \polys{} and a series of competitive baselines for few-shot task adaptation on the T0 task suite \citep{sanh2022multitask} and Super-Natural Instructions~\citep[SuperNI;][]{ni}. Based on our results, we report
% We show that \poly{} consistently outperforms single adapter baselines, across different datasets, and can improve on \poly{}.
that \polys{} outperforms \poly{} and single adapter baselines.  Additionally, we show that, thanks to the increased expressivity of the routing function, it becomes possible to fine-tune only the parameters of the routing function (and not the adapters) during few-shot adaptation: the resulting method, \polys{}-$z$, yields competitive performance while requiring orders of magnitude fewer parameters.

Regarding the second and third questions, we uncover that optimization during multitask pretraining plays a key role in explaining the downstream performance of routing-based PEFT approaches. Specifically, we find that \polys{} exhibits a higher cosine similarity between gradients from different tasks than \poly{} and single-adapter multi-task training. Hence, routing enables more knowledge transfer and less interference across tasks during multi-task pre-training. This finding led us to investigate whether routing is useful also during few-shot adaptation. It has been hypothesized \citep{ponti2022combining} that one of the reasons behind \poly{}'s performance resides in the inductive bias of the modular architecture, which allows test tasks to recombine and locally adapt the most relevant modules. To test this hypothesis, we propose \polysm{}, where the routing function is discarded and all available adapter parameters are averaged before few-shot adaptation. We find that \polysm{} can recover the performance of \polys{}, hinting that \poly{}/\polys{} gains are only a result of better multi-task optimization. Finally, we show that \polysm{} can also be used as an effective zero-shot transfer method by training the average of the pre-trained adapters for a few additional steps on the multi-task training set. This yields gains up to 3\% on absolute accuracy w.r.t. to strong baselines such as T0-11B.

\section{Background}

In cross-task generalization, we are given a set of tasks $\mathcal{T} = \{\mathcal{T}_1, .., \mathcal{T}_{|\mathcal{T}|}\}$, with each task $\mathcal{T}_i$ dataset containing a set of samples  $\mathcal{D}_i = \{(x_1, y_1), ..., (x_n, y_n)\}$. The set of all tasks is partitioned into training and test tasks, $\mathcal{T} = \mathcal{T}_{train} \cup \mathcal{T}_{eval}$, and the objective is to leverage data in $\mathcal{T}_{train}$ and transfer knowledge to facilitate learning of the test tasks $\mathcal{T}_{eval}$. For all the methods discussed, learning takes place in two phases, excluding the original unsupervised pre-training of the language model backbone on a separate corpus. The first phase consists of multi-task pre-training, in which either an adapter, such as LoRA or (IA)$^3$, or the full backbone is trained on the set of training tasks $\mathcal{T}_{train}$. The second phase consists in few-shot adaptation, where the learned adapters are fine-tuned independently on each test task in $\mathcal{T}_{eval}$. We follow the procedure from \citep{t5} and formulate each task as a text-to-text problem, enabling standard maximum-likelihood training with teacher forcing~\citep{bengio2015scheduled} and a cross-entropy loss.

\subsection{Adapters: \texttt{LoRA} \& \texttt{(IA)}$^3$}
\label{sec:lora_ia3}
LoRA~\citep{lora} and (IA)$^3$~\citep{tfew} are two recently proposed adapter architectures that achieve competitive trade-offs between performance and parameter efficiency \citep{mahabadi2021parameter,tfew}. For each linear transformation corresponding to the query ($q$), key ($k$), value ($v$) and output ($o$) of the self-attention layers, LoRA modifies the base model \textit{parameters} as follows:
\begin{equation*}
\tag{LoRA}
\vh^{q,k,v,o} = \mW^{q,k,v,o}_0 x + s\cdot \mA^{q,k,v,o} (\mB^{q,k,v,o})^\top x,
\label{eqn:lora}
\end{equation*}
where $\mW_0$ are the (frozen) weights of the pre-trained model (e.g. T5 \citep{t5}). $\mA, \mB \in \mathbb{R}^{d \times r}$ are low-rank learnable parameters and $s\ge 1$ is a tunable scalar hyperparameter. (IA)$^3$, on the other hand, modifies key and value \textit{representations} in self-attention element-wise, and also modifies the feed-forward MLP ($f$):
\begin{equation*}
\tag{(IA)$^3$}
\vh^{k,v} = \vl^{k,v} \odot (\mW_0^{k,v} x);\,\, \vh^{f} = (\vl^{f} \odot \gamma(\mW_1^{f}x)) \mW_2^{f},
\end{equation*}
where $\vl^{k,v,f} \in \mathbb{R}^d$ are learnable parameters , $\mW_{1,2}^{f}$ the frozen parameters of the feed-forward layer in the backbone, and $\gamma$ a non-linearity. For clarity, we will drop the superscripts $q,k,v,o$ in the rest of the paper.% as the modifications will be the same for each layer.

\subsection{Polytropon: Adapter Routing}
\label{sec:poly}
Typical adapter methods either fully share adapters across tasks or train individual adapters for each task. \poly{} addresses the multi-task problem by softly sharing adapter parameters across tasks. Each \poly{} layer contains 1) an inventory of adapter modules $\mathcal{M} = \{\phi_1, \ldots, \phi_m\}$ with $|\mathcal{M}| \ll |\mathcal{T}|$; 2) a routing function $r(\cdot)$ that chooses which subset of the modules to combine for each task. %Each \poly{} adapter layer contains a fixed set of \textit{modules}: 

Each module corresponds to a LoRA adapter, where $\phi_i$ are its associated parameters $\mA^{(i)}, \mB^{(i)} \in \mathbb{R}^{d \times r}$. $r(\cdot)$ is implemented as a task--module routing matrix $\mZ \in \mathbb{R}^{|\mathcal{T}| \times |\mathcal{M}|}$. $z_{\tau}=\mZ_{\tau, :}\in\mathbb{R}^{|\mathcal{M}|}$ is a routing vector of task ${\mathcal{T}_{\tau}}$, with cell $\emZ_{\tau,j}$ being the probability logits of using module $\phi_j$ for task $\mathcal{T}_{\tau}$ in the current layer. Differently from mixture-of-experts~\citep{fedus2021switch}, which perform token-level top-$k$ routing, $\mZ$ converges to a binary matrix, defining a soft partition over modules. This is achieved by using a Gumbel-sigmoid distribution \citep{jang2016categorical} during training, with $\hat{\emZ}_{\tau,j} \sim \texttt{Gumbel}(\emZ_{\tau,j})$.
At each forward pass, \poly{} can be defined as :
% \vspace{-1pt}
\begin{gather*}
\tag{Poly}
\mA^{\tau} =  \sum_{i} \alpha_i \mA^{(i)};\,
\mB^{\tau} = \sum_{i} \alpha_i \mB^{(i)}
\label{eqn:poly}
\end{gather*}

where $\alpha_i = \frac{\hat{\emZ}_{\tau, i}}{\sum_{j} \hat{\emZ}_{\tau,j}}$ , and $\mA^{(i)},\mB^{(i)},\mA^\tau, \mB^\tau \in \mathbb{R}^{d \times r}$. We normalize the mixing coefficients $\hat{\emZ}_{\tau, i}$ for each task to ensure that the number of active modules does not affect the norm of $\mA^\tau,\mB^\tau$. Overall, this approach enables different subsets of \textit{modules} to be activated for the current layer and combined in a task-specific way. % The derivation using $\texttt{ia3}$ adapters is similar. % The decomposition into different skills allows the model to capture different facets of knowledge.
Following~\ref{eqn:lora}, the output of the \poly{} layer is added to the output of the original layer of the frozen backbone: $\vh = \mW_0 x + s \mA^\tau (\mB^{\tau})^{\top} x.$

During multi-task pre-training, for each query, key, value, and output projection in self-attention layers, the parameters learned by \poly{} are the adapter parameters, $\{\mA_i, \mB_i\}_{i=1}^{|\mathcal{M}|}$, and the routing matrices $\mZ$. During fine-tuning, for each test task $\tau$, \poly{} randomly initialize the routing vector $z_{\tau} \in \mathbb{R}^{1 \times |\mathcal{M}|}$ and fine-tunes both $z_{\tau}$ and all the modules parameters $\mathcal{M}$.

% Our methods focus on \textit{routing} function, demonstrating how systematic generalization can be cultivated through \textit{routing} in multi-task contexts. Section \ref{eqn:polys} introduces \polys{}: a method employing multi-head adapter routing. To assess the importance of learning the \textit{routing} function, we investigate this function from two perspectives: Section \ref{sec:polym} presents \polym{}, a method using a \textit{fixed} routing function that directly averages the adapters. Conversely, Section \ref{sec:polyz} introduces \polyz{}, a strategy that fine-tunes the routing function while keeping the adapters static. 
%\LC{We use the following notation throughout the paper, writing a method $x-y$, as using}

% BEGIN TABLE
% \iffalse
\begin{table*}[t]
\centering
\begin{tabular}{lccc}
\toprule
Method & Pre-Training & Fine-Tuning & Inference \\
\midrule
\texttt{Full FT} & $d \times d$ & $d \times d$ & $d \times d$ \\
\midrule
\texttt{LoRA} & $d \times 2r$ & $d \times 2r$ & $d \times 2r$ \\
\poly{} & $ d \times 2r \times |\mathcal{M}| + |\mathcal{T}| \times |\mathcal{M}|$ & $ d \times 2r \times |\mathcal{M}| + |\mathcal{M}|$ & $ d \times 2r$ \\
\poly{}-$z$ & $ d \times 2r \times |\mathcal{M}| + |\mathcal{T}| \times |\mathcal{M}|$ & $|\mathcal{M}|$ & $|\mathcal{M}|$ \\
\midrule
\polysm{} & $ d \times 2r \times |\mathcal{M}| + |\mathcal{T}| \times |\mathcal{M}|$ & $ d \times 2r$ & $ d \times 2r$ \\
\polys{}-$z$ & $ d \times 2r \times |\mathcal{M}| + |\mathcal{T}| \times |\mathcal{M}| \times h$ & $|\mathcal{M}| \times h$ & $|\mathcal{M}| \times h$\\
\polys{} & $ d \times 2r \times |\mathcal{M}| + |\mathcal{T}| \times |\mathcal{M}| \times h$ & $ d \times 2r \times |\mathcal{M}| + |\mathcal{M}| \times h$ & $ d \times 2r$ \\

\bottomrule
\end{tabular}
\caption{Number of parameters (per layer) used for each method. The calculation uses \texttt{LoRA} as the base adapter, modifying a linear transform in $\mathbb{R}^{d \times d}$. Note that the total number of parameters changed by \texttt{Full FT} is larger, given that the method also changes parameters for layers not modified by \texttt{LoRA}. 
%All of the methods tested (apart from \texttt{Full FT}) use the same number of parameters at inference time.
}
\vspace{-5pt}
\label{tab:ft-param}
\end{table*}

% \fi
% END TABLE 

\section{Multi-Head Adapter Routing (\polys{})}
\label{sec:polys}

In \poly{}, module combination remains \textit{coarse}: only linear combinations of modules are possible, and thus the resulting aggregated adapter remains a linear function of the modules. We propose to augment the expressivity of the module combination while keeping the parameter count similar. \polys{} (Fig.~\ref{fig:first_fig}) takes inspiration from multi-head attention~\citep{transfo}: it partitions the input dimensions into $h$ different disjoint blocks, performs a separate \poly-style combination for each of them, and finally concatenates them. This corresponds to learning a different routing matrix $\mZ$ for each block of input features, therefore enabling the model to select different adapters for different blocks of the input dimensions. This aggregation approach is \textit{piecewise} linear (i.e., linear within disjoint intervals), which allows for more expressive combinations of modules.  

In each \polys{} layer, the routing function is a third-order tensor $\tZ \in \mathbb{R}^{|\mathcal{T}| \times |\mathcal{M}| \times h}$, where $\tZ_{:, :, h}\in\mathbb{R}^{|\mathcal{T}| \times |\mathcal{M}|}$ is a 2D slice of the tensor $\tZ$. A slice represents the routing matrix for each of the $h$ heads. Let us denote with $\mW[k] \in \mathbb{R}^{\frac{d}{h} \times r}$ the $k$-th partition along the rows of the matrix $\mW \in \mathbb{R}^{d \times r}$. The adapter parameters $\mA^\tau \in \mathbb{R}^{d \times r}$ for task $\tau$, and for each adapter layer, are computed as (similarly for $\mB^\tau$):
\begin{gather*}
\tag{MHR}
\mA^\tau_{k} = \sum_{j} \mA_j[k] \cdot \frac{\hat{\etZ}_{\tau, j, k}}{\sum_{j} \hat{\etZ}_{\tau, j, k}}\;\; \textrm{with}\;\; \mA^\tau_{k} \in \mathbb{R}^{\frac{d}{h} \times r}, \\
 \mA^\tau = \texttt{concat}(\mA^\tau_{1}, \ldots, \mA^\tau_{h})
\label{eqn:polys}
\end{gather*}
where \texttt{concat} concatenates along the first dimension. Multi-task pre-training and fine-tuning are similar to \poly{}. Note that \polys{} results in only a negligible increase in the total amount of parameters, since most of the parameters are contained in the LoRA weights $\mA, \mB$ (Tab.~\ref{tab:ft-param}).

\iffalse
\begin{figure}[t]
\centering
% \includegraphics[width=0.9\linewidth]{figs/poly_granularity.png}
\includegraphics[width=0.99\linewidth]{figs/figure_2.png}
\caption{Different ways to control the expressivity of routing-based methods. \textit{Left: } In Polytropon, one can only add additional modules, resulting in a linear parameter increase. \textit{Middle : } In \polys{}, additional heads only introduce routing matrices $Z$, resulting in a negligible parameter increase. \textit{Right : } We also explore tying the module allocation $Z$ weights across adapters in the same layer for more parameter efficient adaptation.} 
\label{fig:poly_granularity_appendix}
\end{figure}
\fi

\paragraph{Routing-Only Fine-Tuning (\polys{}-$z$)}
\label{sec:polyz}
Prior work~\citep[\textit{inter alia}]{shao2023compositional} has shown that compositional generalization can be achieved by learning to (re-)combine in novel ways pre-existing modules. We investigate whether fine-tuning the module parameters is really needed for few-shot adaptation in the context of  both \poly{} and \polys{}. Therefore, we name \poly{}-$z$ and \polys{}-$z$ the variants that, during few-shot adaptation, keep the parameters of the modules learned during multi-task pre-training fixed and just update the routing parameters $\mZ$. Crucially, this enables highly parameter-efficient adaptation: for \texttt{LoRA} adapters, $\mA$ and $\mB$ matrices constitute the overwhelming majority of parameters. Therefore, by freezing the $\mA, \mB$ matrices and only updating $\mZ$, we can significantly reduce the parameter cost when transferring knowledge to a new task.

\paragraph{Adapter Average Fine-Tuning (\polysm{})}
\label{sec:polym}
To assess the importance of the routing parameters during few-shot adaptation, we propose an additional variant of \polys{}, \polysm{}, which shares the same multi-task pre-training procedure as \polys{}, but for each test task $\tau$, fixes $z_{\tau} = (1 / |\mathcal{M}|, \ldots, 1 / |\mathcal{M}|)$ during few-shot adaptation. This is equivalent to discarding the routing parameters and averaging the module parameters into a single one before fine-tuning. Specifically, the adapter used during fine-tuning is initialized with (similarly for $\mB^\tau$):
\begin{gather*}
\tag{\polys{}-$\mu$}
\mA^\tau = \frac{1}{|\mathcal{M}|}\sum_{i} \mA^*_i;\;  \mA^\tau \in \mathbb{R}^{d \times r}
\label{eqn:polym}
\end{gather*}
where $\mA^*_i$ are the parameters of the adapters after \polys{} multi-task pre-training.
Note that, differently from \polys{}, \polysm{} fine-tunes the same amount of parameters as the single adapter baseline. Thus, any difference in performance between the single adapter baseline and \polysm{} comes from differences in the adapter initialization and must be due to the optimization process taking place in the multi-task pre-training, before few-shot adaptation.

%%% IF WE ARE SHORT ON SPACE, we can wrapfig only the right-most part of figure 2
\iffalse
\begin{wrapfigure}[20]{r}{0.5\textwidth}
  \centering
  % \vspace{-10pt}
  \includegraphics[width=0.8\linewidth]{figs/sharing_Z_only.png}
  \captionsetup{width=.9\linewidth}
  \caption{We explore tying the module allocation $Z$ weights across adapters in the same layer for more parameter-efficient adaptation.}
  \label{fig:poly_granularity}
\end{wrapfigure}
\fi
%%%

\label{sec:poly_granularity}
\paragraph{Routing Granularity} 
In the original \poly{}, \cite{ponti2022combining} showed that learning a routing matrix $\mZ$ for each model layer gave better performance than sharing a single $\mZ$ matrix across all layers. 
We therefore investigate whether this holds true also for its multi-head counterpart, \polys{}. In addition, we explore intermediate approaches between one $\mZ$ per layer and a single one shared for the entire model. %As shown in Fig. \ref{fig:poly_granularity},
In particular, we consider sharing $\mZ$ 1) for the adapter layers belonging to the same Transformer block; or 2) for every block of $l$ layers, which enables us to easily trade off expressivity for parameter efficiency. As we will demonstrate in section \ref{sec:results_ztuning}, this is an efficient mechanism to navigate this Pareto front in regimes of very small budgets of parameters per task.

\section{Experiments}
Our experimental evaluation aims to answer three research questions: 1) Does the expressivity of the routing function
matter? 2) Why do routing-based PEFT methods yield superior performance? 3) Is routing
useful during both multi-task pre-training and few-shot adaptation? We first present the baselines and datasets used in our evaluation and then discuss each question in turn.\footnote{
We note that all experiments were run on a single NVIDIA A100 GPU.} 

\subsection{Baselines}
In addition to~\poly{}, we compare \polys{} to the following baselines for task-level generalization.

\textbf{\texttt{LoRA}}/\textbf{\texttt{(IA)$^3$}} trains a single adapter common to all pre-training tasks, which is then fine-tuned on each test task separately. This is arguably the most widespread approach for parameter-efficient cross-task generalization~\citep{tfew,pfeiffer2023modular}.

\label{sec:as}
\textbf{\AS{}} \cite{chronopoulou2023adaptersoup} trains a different adapter for each task. The method only averages the adapter weights of the training tasks most similar to a given test task, before proceeding with few-shot adaptation. To compute task relatedness, we measure the cosine similarity of sentence embeddings for each task averaged over their training dataset. Notably, unlike the methods proposed in this paper, there is no knowledge sharing (nor interference) during multi-task pre-training as task adapters are trained independently.

% \textbf{\AS{}} \cite{chronopoulou2023adaptersoup} trains a different adapter for each task or domain, sharing the same pretraining step as \private{}. However, rather than a uniform weight average, the method only averages the adapter weights of the training tasks most similar to a given test task, before proceeding with few-shot adaptation. \LC{To compute task relatedness, we measure the cosine similarity of sentence embeddings for each task averaged over their training dataset.}

% \item\fullft{} -- for benchmarks where natural language instructions for each task are available, we also experiment with fully fine-tuning a model. In this case, no parameters of the pre-trained model are frozen and no adapters are inserted. Note that this increases the amount of trainable parameters significantly compared to all the other baselines. The purpose of this baseline is to verify if modular architectures are necessary when natural language instructions already provide information to condition the model behaviour.

\subsection{Datasets}
We test our methods on the T0~\cite{sanh2022multitask} evaluation suite, following the same setup as \cite{tfew}, and \texttt{SuperNI}~\cite{ni}, a large-scale dataset with more than 1,600 training tasks. 
% ; and Crossfit~\cite{xfit} a multi-task dataset with 160 train and 20 test tasks originally used to benchmark Polytropon. We focus the core of our analysis on 

\paragraph{T0 Tasks} We follow the pre-training and fine-tuning procedure discussed in \cite{tfew}, using hyper-parameters and losses suggested in the public codebase for T-Few.\footnote{\url{https://github.com/r-three/t-few}} 

All methods were tested with T5-XL~\cite{t5} and T0-3B~\cite{sanh2022multitask} as the backbone model. Crucially, T5 is simply pre-trained on (masked) language modelling, whereas T0 is further instruction tuned: in particular, the full model is fine-tuned on examples from multiple training tasks that have been augmented with task instructions. To ensure fairness for all methods, we report the median and standard deviation of the best validation accuracy for each test task across 3 seeds, when evaluated every 50 training epochs. We treat each data subset--template pair as a unique task, yielding a total of 313 tasks.

\paragraph{SuperNI} To limit computational costs, we report the result on 20 out of 119 test tasks. Tasks were chosen at random, with the requirement that at least 300 examples were available, and were equally split into 100 training, 100 validation and 100 test examples. For every method, we perform early stopping on the validation set. We report results with Rouge-L averaged across 3 seeds. All methods use T5-XL~\citep{t5} as the backbone and not T0, as T0 training tasks and SuperNI test tasks may overlap. % \LC{TODO : modify with 3B}

%\paragraph{Crossfit} For the Crossfit \cite{xfit} experiments, we use the same hyperparameters as in the original Polytropon paper for both pre-training and fine-tuning. In order to perform early stopping, we use perplexity on the validation sets of the training tasks, rather than performing few-shot fine-tuning on the validation tasks. We were able to reproduce (and in some cases increase) the performance of \poly{} and \shared{}, the two methods tested in the original paper. All methods use BART-large~\cite{lewis-etal-2020-bart} as the backbone and LoRA~\cite{lora} for our parameter-efficient adapters. 

% Note that we reran the \tfew{} baseline since the original weights were obtained with mixed-precision training. 

% Finally, we note the fine-tuning procedure consists of 1000 updates on the training data, without any cross-validation. This prohibits us from tuning the remaining \poly{} hyper-parameters. 

% \input{xfit_bigtable}

\section{Results and Discussion}
% \input{tables/t0_table}
% \input{tables/larger_t0_table}
% \input{tables/ni_table}
% \input{tables/ale_table}
% In this section, we present and analyse the results over the aforemented benchmarks and several parameter budgets. 
% In this section, we provide a comprehensive analysis of our proposed approach, highlighting its performance across different settings and its superiority in terms of parameter efficiency.

\begin{figure}[t]

\begin{minipage}{0.5\textwidth}
\vspace{10pt}
\centering
\begin{tabular}{lccc}
\toprule
\textbf{T0 Dataset} &  \textbf{Avg. Test} \\
\midrule
\multicolumn{2}{l}{\emph{Backbone T5-XL-LM}} \\
% \texttt{Multi-Task Full Finetuning + LoRA} & 68.9$_{x.x}$ \\
\texttt{(IA)}$^3$ & 62.4$_{0.4}$ \\
\AS{} & 62.1$_{1.0}$ \\
\texttt{LoRA}  & 66.0$_{1.6}$ \\
\texttt{LoRA-big} & 65.4$_{0.9}$  \\ 
\poly{}-$z$ & 66.4$_{0.3}
$ \\
\poly{} & 68.0$_{1.0}$ \\
\rowcolor{lightgray}\polys{}-$z$  & 68.3$_{0.8}$ \\
\rowcolor{lightgray}\polys{}  & \underline{69.1}$_{1.0}$ \\
\midrule
\multicolumn{2}{l}{\emph{Backbone T0-3B}} \\
\tfew~\cite{tfew} &  66.2$_{0.5}$ \\
\AS{} &66.1$_{0.6}$ \\
\texttt{LoRA}  & 67.4$_{0.8}$ \\
% \texttt{LoRA-big} &  68.0$_{0.8}$  \\ 
\poly{}-$z$ &  65.3$_{1.0}$  \\
\poly{} &  69.0$_{0.8}$ \\
\rowcolor{lightgray}\polys{}-$z$ & 68.4$_{1.2}$  \\
\rowcolor{lightgray}\polys{}  &\underline{69.3}$_{1.2}$ \\
\bottomrule
\end{tabular}
% \captionsetup{width=1.2\linewidth}
% \captionof{table}{(top) Results on T0 dataset \cite{sanh2022multitask}, we report the mean of the best validation accuracy for each test task. (bottom) Results on SuperNatural Instructions dataset. Subscripts are interquantile range.}
\hspace{0.01\textwidth}
\end{minipage}%
\begin{minipage}{0.55\textwidth}
\centering
\vspace{0pt}
\includegraphics[width=\textwidth]{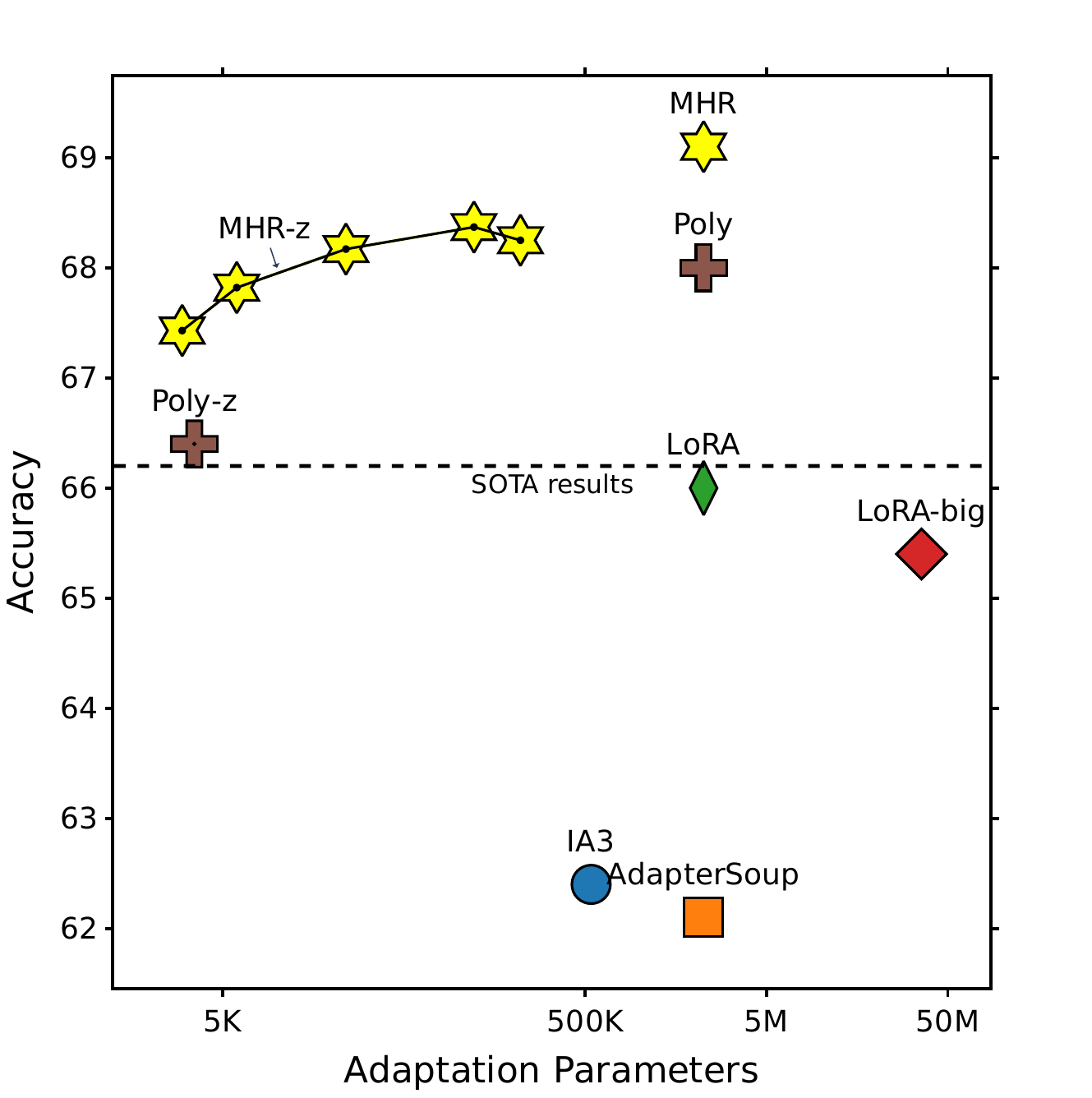}
% \captionsetup{width=.75\linewidth}
% \caption{Accuracy of PEFT methods when applied to T5-XL. The x-axis shows the parameter count during the fine-tuning process.} 
\end{minipage}
\caption{\textit{Left:} Results of few-shot adaptation on T0 dataset \cite{sanh2022multitask}. We report the mean of the best validation accuracy for each test task. Subscripts correspond to standard deviation. \textit{Right:} Accuracy of PEFT methods on the T0 dataset when applied on top of T5-XL. The x-axis shows the parameter count during the fine-tuning process.}
\label{fig:main_results}
\end{figure}

\subsection{Does the expressivity of the routing function matter?}

\paragraph{\polys{} outperforms PEFT approaches}
We start our analysis by evaluating the effectiveness of our proposed technique when applied over a backbone that has not undergone prior training on instruction-following data (T5-XL). As indicated in the T0 benchmark results in the top table of Fig.~\ref{fig:main_results}, it is clear that multi-head routing techniques have a distinct advantage, outperforming both single-head routing \poly{} by \textbf{1.1\%}, and surpassing standard \shared{} approaches by an impressive \textbf{3.1\%}. We also study the impact of performing instruction tuning of the full backbone before adapter training. To this end, we also experiment with T0-3B as a backbone. In the bottom table of Fig.~\ref{fig:main_results}, we can observe that while the relative gap between \polys{} and baselines is smaller, multi-head routing still manages to yield favourable results. Hence, the gains of \polys{} compound with other multi-task methods such as instruction tuning.
% For completeness, in appendix \ref{tab:appendix_all_results}, we report additional results over this backbone, and observe similar results.
%Subsequently, we align our approach more closely with the methodology employed in \tfew{} \cite{tfew}, wherein we utilize an instruction-tuned backbone \texttt{T0-3B}. For fairer comparison with \tfew{}, in this version, we patch the same layers as~\tfew{},~i.e. only the \texttt{k,v,ff} layers (refer to \ref{sec:lora_ia3}). This again yields similar results, with \polys{} demonstrating considerable improvements over the existing models. 
Finally, we turn our attention towards the SuperNI dataset (Tab.  \ref{tab:ni_main}). Here, \polys{} continues to surpass analogous baselines.

\label{sec:results_ztuning}
\paragraph{\polys{}-$z$ facilitates extreme parameter efficiency} Fig.~\ref{fig:main_results} (right) reveals intriguing findings regarding \polys{}-$z$. When we restrict training to only the routing parameters $\mZ$ in the original \poly{}, the results are unfortunately not up to par with its version where both routing and adapters are updated. However, when we apply the same constraint to \polys{}, the performance is significantly closer to the optimum achieved in this setting. In fact, \texttt{MHR}-$z$ surpasses prior baselines while simultaneously necessitating fewer parameters for effective adaptation to new tasks. 
%. 
% A more in-depth analysis of how the granularity of the routing matrix $Z$ balances performance and efficiency can be found in the Appendix \ref{sec:poly_granularity_appendix}. 
Moreover, by controlling the number of layers which share the same $Z$ allocation (see sec. \ref{sec:poly_granularity}), \texttt{MHR}-$z$ is able to trade-off performance for parameter efficiency, even surpassing \poly{}-$z$ in settings with only 3K trainable parameters per test task 
 (see also $\S$ \ref{sec:poly_granularity_appendix} for a more in-depth analysis).
This trend is similarly observed in the SuperNI benchmark (Tab.  \ref{tab:ni_main}), where updates restricted to the routing parameters yield performance on par with standard fine-tuning. We therefore conclude that the \texttt{MHR}-$z$ represents a robust approach for achieving extreme parameter efficiency in adaptation. 

\begin{wraptable}{r}{0.5\textwidth}
\centering
\begin{tabular}{lc}
\toprule
\textbf{SuperNI Dataset} & \textbf{Rouge-L} \\
\midrule
%\texttt{Full-FT} + \texttt{(IA)}$^3$ & 68.0$_{0.8}$ \\
%\texttt{Full-FT} + \texttt{LoRA} & 67.9$_{0.7}$ \\
\texttt{LoRA}  &  67.6$_{0.8}$ \\
\texttt{LoRA-big} & 67.2$_{0.7}$ \\ 
\poly{}-$z$ &  64.6$_{0.3}$\\
\poly{} &  67.8$_{0.8}$\\
\rowcolor{lightgray}\polys{}-$z$   & 68.0$_{0.2}$\\
\rowcolor{lightgray}\polys{} & \underline{68.5}$_{0.3}$ \\ 
\bottomrule
\end{tabular}
\captionsetup{width=.9\linewidth}
\caption{Results on \textbf{SuperNI} dataset. Subscripts are standard deviation.}
\label{tab:ni_main}
\end{wraptable}

\paragraph{Additional routing heads is more beneficial than extra modules}
\iffalse
\begin{wrapfigure}[17]{r}
{0.5\textwidth}
  \centering
  \includegraphics[width=0.99\linewidth]{figs/skill_vs_split.pdf}
  \captionsetup{width=.9\linewidth}
  \caption{Comparing an increase of routing heads with additional modules.}
  \label{fig:skill vs split}
\end{wrapfigure}
\fi
In the original \poly{} approach, a tradeoff between capacity and parameter efficiency can be achieved by adding extra modules for each adapter layer. However, this results in a linear increase in the number of multi-task parameters, which can become impractical. To explore a more effective tradeoff, we investigate the option of adding additional routing heads instead of extra modules. Fig \ref{fig:grad_alignment_and_skills_vs_split} (right) presents the comparison between the two approaches. It demonstrates that increasing the number of routing heads leads to better performance compared to adding more modules. Importantly, the benefit of multi-head routing is twofold: it provides increased expressivity for the model, while also maintaining parameter efficiency. This finding highlights the advantage of multi-head routing as a more effective approach for balancing expressivity and parameter count in few-shot adaptation scenarios.

% \vspace{-5pt}
\begin{wraptable}[13]{r}{0.4\textwidth}
\centering

\begin{tabular}{lccc}
\toprule
\textbf{T0 Dataset} &  \textbf{Avg. Test} \\
\midrule
%\multicolumn{2}{l}{\emph{Backbone T5-XXL}} \\
%\texttt{LoRA}  & 70.1$_{0.7}$ \\
%\poly{}-$z$ & 71.6$_{0.8}$ \\
%\poly{} & \underline{74.0}$_{0.9}$ \\
%\rowcolor{lightgray}\polys{}-$z$  & 73.3$_{1.0}$ \\
%\rowcolor{lightgray}\polys{}  & 73.9$_{0.8}$ \\
%\midrule
\multicolumn{2}{l}{\emph{Backbone T0-11B}} \\
\tfew \ \cite{tfew}&  72.5$_{0.9}$ \\
\texttt{LoRA}  & 72.3$_{1.0}$ \\
\poly{}-$z$ &  70.0$_{0.6}$  \\
\poly{} &  \underline{74.9}$_{0.6}$ \\
\rowcolor{lightgray}\polys{}-$z$ & 72.9$_{0.8}$  \\
\rowcolor{lightgray}\polys{}  &74.7$_{0.6}$ \\
\bottomrule

\end{tabular}

\caption{Few-shot results over 11B parameter backbones.}
\label{tab:11b}
\end{wraptable}

\paragraph{Routing-based methods also excel at the 11B scale} \textcolor{black}{We proceed to evaluate if \poly{} and \polys{} surpass established PEFT approaches when trained over a larger model backbone. To accomplish this, we employ the 11B version of T0. As depicted in Tab. \ref{tab:11b}, routing-based methods once again outshine standard adapter training, surpassing our reproduction of the previous state-of-the-art in \cite{tfew} by over 2\%. We observe that \poly{} and \polys{} show similar performance in standard fine-tuning, but \polys{} $z$-tuning remains more performant in routing-only fine-tuning. Indeed, \polys{}-$z$ (221K params) outperforms \poly{}-$z$ (3.5K params) by 2.9\%, while still remaining more parameter efficient than \cite{tfew} (1.1M params).  % Intriguingly, $z$-tuning methods demonstrate better performance on the T5 backbone compared to T0, which is not the case for standard fine-tuning. This indicates that at a larger scale, fine-tuning the backbone may negatively impact later modularization. 
}

% Specifically, using 64 heads with the $\texttt{[blockwise]}$ discussed in section \ref{sec:poly_granularity}, the model can outperform standard LoRA adapters and \poly{}, using only 25K params (or 0.0008\% of the backbone). 

% \paragraph{Multi-head routing is a better use of $Z$ parameter budget} \LC{TODO, not sure if this will stay for now. I wan to show that you are better off using e.g. 16 heads with $\texttt{blockwise}$ than standard fine-grained poly-S, while both have a similar amount of $Z$ params.}

\subsection{Why do routing-based PEFT methods yield superior performance?}

While our proposed methods have demonstrated promising results across a broad spectrum of datasets and varying adaptation parameter budgets, the question of \textit{why} routing-based PEFT exhibits superior performance remains unanswered. In this section, we aim to uncover the key components that drive \polys{}'s superior performance.

\paragraph{Learning the Routing Function is essential} 
Given that \poly{} and \polys{} have access to more parameters than standard adapters during multi-task pretraining, we investigate whether this, and not the routing mechanism, is responsible for their superior performance. To do so, we compare them to a baseline approach. 
Instead of learning the routing function, we randomly assign a binary module allocation to each data point in a minibatch, disregarding task information. This random routing approach, akin to \cite{wang2022adamix}, allows us to directly assess the influence of additional parameters during multi-task training. At test time, the learned modules are averaged into a single one before fine-tuning; we therefore refer to this baseline as \random{}. 
% As shown in Table \ref{tab:ablate_one}, (This table is not in the paper)
On the T0 benchmark with the T5-XL backbone, \random{} performs similarly to a standard LoRA adapter (66.0\%), while \poly{} and \polys{} outperform it by\textbf{ 2\%} and \textbf{3.1\%} respectively. Therefore, we conclude that learning a routing function is crucial, and merely increasing capacity during training does not directly lead to improvements.

\begin{figure}[t]
\centering
\includegraphics[width=0.46\linewidth]{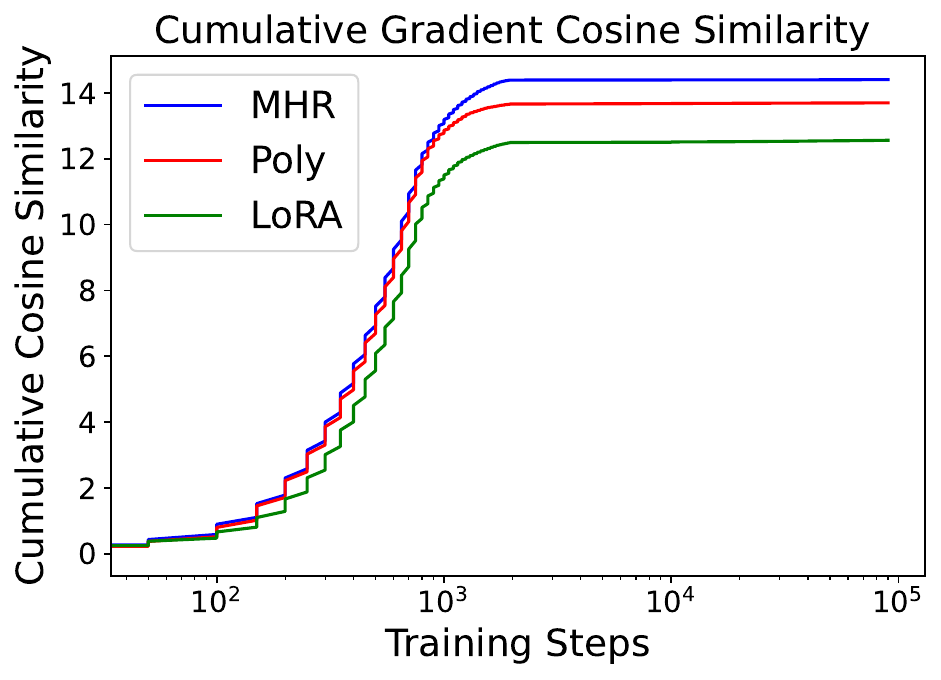}
\includegraphics[width=0.51\linewidth]{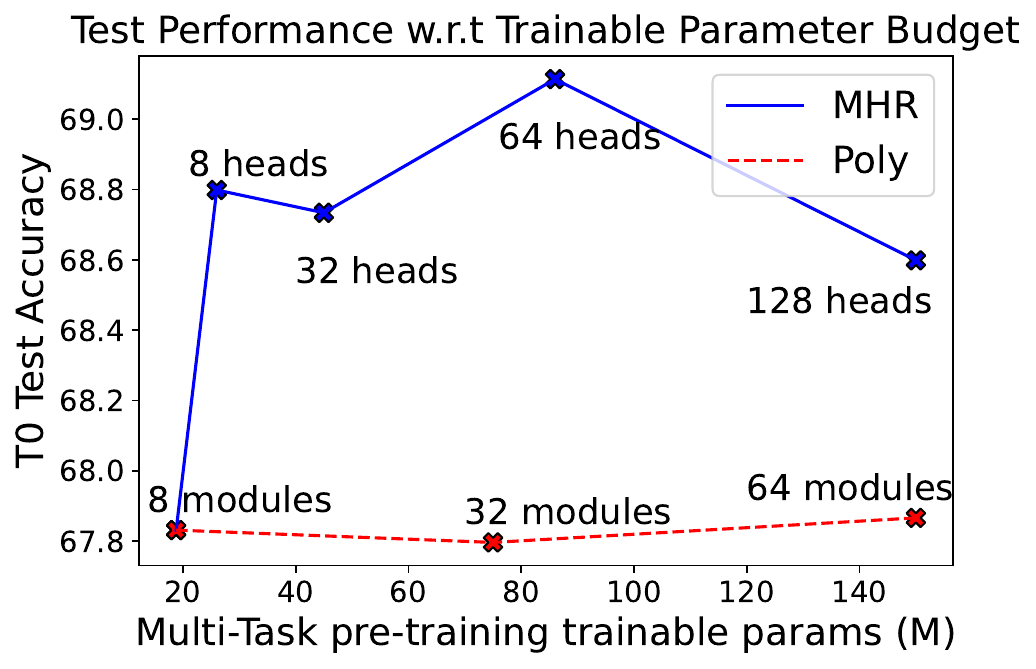}
\caption{\textit{Left:} Gradient alignment between tasks during multi-task pretraining. \textit{Right:} Increasing the number of heads offer better scaling properties than increasing the number of modules.}
\label{fig:grad_alignment_and_skills_vs_split}
\end{figure}

\paragraph{\polys{} fosters transfer and mitigates interference across pretraining tasks} Recognizing the pivotal role of the multi-task pretraining step in bolstering \poly{}'s performance, we explore the extent of transfer and interference across training tasks. By monitoring the average gradient alignment for each task pair (in terms of cosine similarity) throughout the training process, we are able to gauge the level of positive transfer. As Fig.~\ref{fig:grad_alignment_and_skills_vs_split} (left) shows,  \polys{} displays a greater degree of gradient cosine similarity across tasks compared to other PEFT alternatives, including \poly{}. This finding suggests that the enhanced flexibility offered by multi-head routing may serve to mitigate interference across tasks to a larger extent than standard routing while simultaneously promoting positive transfer.

\begin{wraptable}[14]{r}{0.45\textwidth}
\centering
\vspace{-30pt}
\begin{tabular}{lc}
\toprule
\textbf{T0 Dataset} & \textbf{Test Acc.} \\
\midrule
\texttt{LoRA}  & 66.0$_{1.6}$ \\
\texttt{AdapterSoup} & 62.1$_{1.0}$ \\
\poly{} & 68.0$_{0.8}$ \\
\polym{}  & 67.8$_{0.6}
$ \\
\rowcolor{lightgray}\polys{} & 69.1$_{1.1}$ \\
\rowcolor{lightgray}\polysm{}  & 69.1$_{0.9}$ \\
\midrule
\textbf{SuperNI} & \textbf{Rouge-L} \\
\midrule
\texttt{LoRA}  &  67.6$_{0.8}$ \\
\poly{} & 67.8$_{0.8}$ \\
\polym{}  & 68.3$_{0.5}$ \\
\rowcolor{lightgray}\polys{}& 68.5$_{0.6}$ \\
\rowcolor{lightgray}\polysm{}  & 68.5$_{0.8}$ \\
\bottomrule
\end{tabular}
\captionsetup{width=\linewidth}
\caption{Evaluating the impact of modular adaptation at test time.}
\label{tab:poly_mu}

\end{wraptable}

% \paragraph{The significance of transfer during Multi-Task training} 
% \input{tables/ablation_table}
% Interference across training tasks has been postulated as a significant impediment during multi-task optimization \cite{wang2021gradient}. The following investigation focuses on discerning whether prevention of interference, or the promotion of transfer, has a more substantial impact on downstream performance. Table \ref{tab:ablate_one} presents a comparison of the performance of a single \shared{} adapter across tasks, juxtaposed with several variants that train independent adapters for each training task. Namely, in addition to the  \AS{} method described above, we also evaluatie the \AS$-\mu$ variant, which uniformly averages all adapter weights (as in \polym{}). In these baselines, we employ a \textit{coarse} task definition of one for each dataset, producing 38 tasks, or a task for each dataset-template pair, resulting in 313 \textit{refined} tasks. The findings suggest that the baselines that share more parameters across tasks generally demonstrate superior performance, thereby implying that transfer is indeed of more importance. %We also note that averaging over a subset of related training tasks (top k=5) is a better weight initialization than considering a single (k=1) or all (\AS$-\mu$) tasks. This suggests that 

% \input{tables/ablate_and_polymu}

\subsection{Is routing important for task generalization?}
We assessed the importance of routing during pre-training. We now proceed to verify whether it is important to learn routing during few-shot adaptation, too. \polym{} and \polysm{} consistently outperform \shared{}, and match the performance of \poly{} / \polys{} (Tab.~\ref{tab:poly_mu}). This demonstrates that, for few-shot adaptation, the average of the pre-trained modules provides a better initialization than learning an adapter shared across all the tasks during pre-training. The consistently superior performance of \polym{} with respect to \random{} and \AS{} stresses the importance of routing during multi-task pre-training (but not during adaptation), which provides an effective adapter initialization for few-shot learning. This finding could potentially inspire future work for improving meta-learning and weight-averaging approaches~\citep{DBLP:journals/corr/abs-1803-05407}.

\vspace{10pt}
\paragraph{\polysm{} excels at zero-shot learning} For many downstream tasks of interest, additional labelled data may not be available. In such settings, it is unclear how to leverage \polysm{} and \polym{} methods.
To address this, we fine-tune the average of the multi-task trained adapters on the multi-task pre-training data (instead of using the downstream few-shot data), for an additional $k$ steps. The results are presented in Table~\ref{tab:0shot}. We find that without any additional fine-tuning ($k=0$), averaging the adapters does not yield good results. This is due to a potential mismatch between adapters learned via task-specific routing, and the uniform routing strategy. We can observe that when fine-tuning the average of the adapters on the multi-task pre-training data for an additional $k$ steps, \polysm{} show strong performance when evaluated in a zero-shot manner. For a fair comparison, we also additionally fine-tune \shared{} for the same number of additional steps. Our best model achieves a zero-shot performance of 64.5 on top of T0-11B, achieving an absolute gain of 3.5\% accuracy points.

\begin{table}
\centering
\begin{tabular}{lcccc}
\toprule
\textbf{Method} & \multicolumn{4}{c}{\textbf{Zero-Shot Test with \textit{k}-shot Extra Training}} \\
 & $k=0$ & $k=1 000$ & $k=5 000$ & $k=10 000$ \\
\midrule
{\emph{Backbone T5-XL-LM}}  & 43.2&&\\
% \texttt{Multi-Task Full Finetuning + LoRA} & 68.9$_{x.x}$ \\

\texttt{LoRA}                 & 56.5 & 56.0 & 56.1 & 55.7 \\
\polym{}                       & 46.0 & 53.0 & 56.8 & 56.3 \\
\rowcolor{lightgray}\polysm{}  & 48.0 & \underline{58.0} & 57.1 & 56.3 \\
\midrule
{\emph{Backbone T0-11B}}\,\citep{sanh2022multitask} & 61.0 &&& \\
\texttt{LoRA}                 & 61.2 & 61.6 & 61.5 & 61.5 \\
\polym{}                       & 62.1 & 63.6 & 63.9 & 64.4 \\
\rowcolor{lightgray}\polysm{}  & 63.5 & \underline{64.5} & 64.5 & 64.4 \\
\bottomrule
\end{tabular}
\vspace{7pt}
\caption{Zero-shot performance for MHR and the baselines, reported as the average over the 11 evaluation datasets from \cite{sanh2022multitask}. To obtain these zero-shot results, we average the learnt \poly{}/\polys{} adapters before performing $k$ additional fine-tuning steps on the multi-task pretraining data. This effectively enables zero-shot transfer to downstream tasks using the same amount of parameters/flops as the baseline LoRA. \polys{} outperform baseline LoRA by up to 3\% absolute accuracy points on T0-11B.}
\label{tab:0shot}
\end{table}

\section{Related Work}
Multi-task learning is effective for low-resource tasks~\citep{wei2021finetuned,aribandi2021ext5,sanh2022multitask}, as knowledge can be borrowed from similar tasks by sharing the model parameters. Multi-task learning has also been applied across languages and modalities~\citep{ponti2019modeling,bugliarello2022iglue}. In the context of NLP, several families of methods enable learning new tasks from a limited set of labelled examples. Few-shot in-context learning \citep[ICL;][]{gpt3}, where examples of a new task are concatenated into an input prompt, enables models to generalize to \textit{unseen} tasks without any gradient-based training. Such approaches are however sensitive to the prompt format and example ordering \citep{zhao2021calibrate}. More importantly, ICL methods incur a significant compute overhead, as for every prediction, the full set of examples must be processed by the model \citep{tfew}. To remedy this, many parameter-efficient fine-tuning (PEFT) methods have been proposed as an alternative to ICL, where a small number of new parameters are added over the frozen pretrained network. To name a few, LoRA \citep{lora} injects learnable low-rank matrices into each Transformer layer. Alternatively, the learnable matrix can be sparse, selecting nonzero shifts via the Lottery-Ticket hypothesis \citep{ansell-etal-2021-mad-g} or via their approximate Fisher information \citep{sung2021training}. Finally, prefix-tuning methods \citep{li-liang-2021-prefix} prepend learnable embeddings to the input or intermediate representations to specialize the model towards a downstream task.

Modular networks partition their parameters into several expert modules, each of them specialized to handle specific sub-tasks \citep{jacobs1991adaptive, kirsch2018modular}. Modular networks are an appealing solution to the problem of adapting to unseen tasks \citep{corona2020modularity}, as the model can leverage its existing modules and recombine them in a novel way, thus achieving systematic generalization \citep{bahdanau2018systematic}. They have also been tested in learning scenarios with data presented sequentially \citep{ostapenko2021continual}, and with changing environments \cite{goyal2019recurrent}. In NLP, mixture-of-experts (MoE) models \citep{shazeer2017outrageously, fedus2021switch}, where a learned gating mechanism routes token representations to appropriate experts (Feed-Forward layers), have shown success in scaling the number of parameters while retaining time efficiency. This results in higher performance when compared to their dense counterparts using a similar compute budget.

\section{Conclusions}
In this paper, we tackle the challenge of generalizing to new tasks based on a few examples after multi-task pre-training. Specifically, we focus on Polytropon \citep{ponti2022combining}, a model where each task is associated with a subset of adapters by a routing function. We investigate how varying the level of control afforded by the routing function impacts performance on two comprehensive benchmarks for multi-task learning, T0 and Super-Natural Instructions. First, a newly proposed variant of the routing function, where multiple heads are responsible for different blocks of input dimensions, improves consistently over all other baselines, including LoRA and (IA)$^3$ adapters.  %Importantly, this family of modular neural architectures is superior to instruction tuning, i.e.\ fully fine-tuning a model on tasks associated with natural language instructions. 
Second, we identify the cause of the success of routing in its ability to prevent interference between tasks, as it yields a better alignment between their gradients.
Third, we find that simple averaging of all multi-task pre-trained adapters before few-shot adaptation to new tasks provides comparable performance, thus offering state-of-the-art performance for single-adapter few-shot learning.
Multi-head routing demonstrates the importance of fine-grained adapter selection for sample-efficient generalization and holds promise to improve other modular methods, such as Mixtures of Experts \citep[MoEs;][]{fedus2021switch} in future research.

% \section*{Acknowledgments}
% Nicolas Le Roux is supported by a CIFAR AI Chair. Additionally, we wish to thank Jonathan Pilault \LC{and Lucas Liu} for their useful comments \LC{on earlier versions of this paper}.

\bibliographystyle{abbrvnat}
\bibliography{ref}
%%%%%%%%%%%%%%%%%%%%%%%%%%%%%%%%%%%%%%%%%%%%%%%%%%%%%%%%%%%%

\appendix
\newpage

\section{Appendix}
\subsection{Additional Results}
More detailed numbers on the T0 \cite{sanh2022multitask} and SuperNI \cite{ni} datasets using different backbones, and different adapter layouts over the base model are found in Table \ref{tab:appendix_all_results}. 
\begin{table}[h]
\small
\centering
\begin{tabular}{lccc}
\toprule
\textbf{Model} &  Multi-Task Params & Adaptation Params & Avg. Test \\
\midrule
\multicolumn{4}{l}{\textbf{T0 Dataset}}
\\
\midrule
\multicolumn{4}{l}{\emph{Backbone T5-XL-LM}} \\
\texttt{Multi-Task Full Finetuning + LoRA} & 2.8B & 2.2M & 68.9$_{x.x}$ \\
\texttt{(IA)}$^3$ & 540K & 540K & 62.4$_{0.4}$ \\
\AS{} & 84M & 2.2M & 62.1$_{1.0}$ \\
\texttt{LoRA}  & 2.2M & 2.2M & 66.0$_{1.6}$ \\
\texttt{LoRA-big} & 35M  & 35M  & 65.4$_{0.9}$  \\ 
\poly{}-$z$ & 17M & 3.5K & 66.4$_{0.3}
$ \\
\poly{} & 17M & 2.2M & 68.0$_{1.0}$ \\
\rowcolor{lightgray}\polys{}-$z$ \emph{(64 h)} & 17M & 220K & 68.3$_{0.8}$ \\
\rowcolor{lightgray}\polys{} \emph{(64 h)} & 17M & 2.2M & \underline{69.1}$_{1.0}$ \\
\midrule
\multicolumn{4}{l}{\emph{Backbone T0-3B}} \\
\tfew~\cite{tfew} & 540K & 540K & 66.2$_{0.5}$ \\
\AS{} & 84M & 2.2M & 66.1$_{0.6}$ \\
\texttt{LoRA}  & 2.2M & 2.2M & 67.4$_{0.8}$ \\
\texttt{LoRA-big} & 35M  & 35M  &  68.0$_{0.8}$   \\ 
\poly{}-$z$ & 17M & 3.5K & 65.3$_{1.0}$  \\
\poly{} & 17M & 2.2M & 69.0$_{0.8}$ \\
\rowcolor{lightgray}\polys{}$z$ \emph{(64 h)}  & 17M & 220K & 68.4$_{1.2}$  \\
\rowcolor{lightgray}\polys{} \emph{(8 h)} & 17M & 2.2M & \underline{69.3}$_{1.2}$ \\
\midrule
\multicolumn{4}{l}{\emph{Backbone T0-3B} \emph{light} version :  (\texttt{k, v, ff} layers only)  }\\
\emph{l}-\texttt{LoRA}~\emph{(rank 1)} & 934K  & 934K & 66.2$_{0.9}$  \\
\emph{l}-\texttt{LoRA}~\emph{(rank 16)} & 15M  & 15M  & 67.6$_{1.1}$  \\
\AS{} (\emph{l}-\texttt{LoRA})          & 35M  & 934K & 64.9$_{1.0}$ \\
\emph{l}-\poly{}-$z$ & 7.5M & 2.1K & 62.9$_{1.2}$  \\
\emph{l}-\poly{} & 7.5M & 934K & 68.0$_{0.5}$ \\
\rowcolor{lightgray}\emph{l}-\polys{}$z$ \emph{(32 h)}  & 7.5M & 74K & 66.8$_{1.1}$  \\
\rowcolor{lightgray}\emph{l}-\polys{}  \emph{(8 h)} & 7.5M & 934K & \underline{68.5}$_{0.7}$ \\ 
% \emph{l}-\polys{} \emph{(64 h)} & 7.5M & 934K & 67.6$_{0.9}$ \\

% model        val   val_std       test  test_std         zs
% 0         few-shot-Z-poly-t03B-lora_p-8x1  61.997729  1.154998  61.997729  1.154998  54.787843
% 5         few-shot-Z-poly-t03B-lora_p-8x8  65.521918  0.616677  65.521918  0.616677  56.154239
% 2  few-shot-Z-poly-t03B-lora_p-8x64-block  66.400081  0.503370  66.400081  0.503370  56.093407
% 4        few-shot-Z-poly-t03B-lora_p-8x64  66.657042  1.278743  66.657042  1.278743  56.153724
% 1        few-shot-Z-poly-t03B-lora_p-8x32  66.837283  1.095762  66.837283  1.095762  55.443070
% 3  few-shot-Z-poly-t03B-lora_p-8x64-layer  67.073684  0.966195  67.073684  0.966195  55.631585

\midrule
\midrule

\multicolumn{3}{l}{\textbf{SuperNI Dataset}} & Rouge-L \\
\midrule
% \multicolumn{4}{l}{\emph{Backbone T5-XL-LM}} \\
% \texttt{LoRA}  & 2.2M & 2.2M &  \\
% \texttt{LoRA-big}
%  & 35M  & 35M & \\ 
% \poly{}-$z$ & 17M & 3.5K & \\
% \poly{} & 17M & 2.2M & \\
% \rowcolor{lightgray}\polys{}$z$ & 17M & 220K & \\
% \rowcolor{lightgray}\polys{} & 17M & 2.2M &  \\
% \midrule
\multicolumn{4}{l}{\emph{Backbone T5-XL-LM} \emph{light} version :  (\texttt{k, v, ff} layers only)  } \\
\emph{l}-\texttt{LoRA}  & 934K & 934K & 67.6$_{0.8}$ \\
\emph{l}-\texttt{LoRA-big} & 18M  & 18M & 67.2$_{0.7}$ \\ 
\emph{l}-\poly{}-$z$ & 7.5M & 2.1K & 64.6$_{0.3}$\\
\emph{l}-\poly{} & 7.5M & 934K & 67.8$_{0.8}$\\
\rowcolor{lightgray}\emph{l}-\polys{}$z$ \emph{(64 h)}  & 7.5M & 147K & 68.0$_{0.2}$\\
\rowcolor{lightgray}\emph{l}-\polys{} \emph{(8 h)}  & 7.5M & 934K & \underline{68.5}$_{0.3}$ \\

\bottomrule

\end{tabular}
\vspace{10pt}
\caption{(top) Results on T0 dataset \cite{sanh2022multitask}, we report the mean of the best validation accuracy for each test task, when evaluated every 50 train epochs. \tfew{} is our reproduction of the results in~\cite{tfew}. \texttt{LoRA-big} means a \texttt{LoRA} adapter with a larger rank. (bottom) Results on SuperNatural Instructions dataset.}
\label{tab:appendix_all_results}
\hspace{0.01\textwidth}

\end{table}

$\texttt{Multi-Task params}$ is the number of additional parameters that must be conserved after multi-task pretraining to enable transfer to a downstream task. $\texttt{Adaptation Params}$ refer to the number of parameters required to learn a new downstream task. For e.g. \poly{} and \polys{}, the multi-task parameters includes the learned modules, but not the  routing  over the training tasks, as these are not required for transfer on a new task. Moreover, variants which average the learned modules prior to fine-tuning (\polys{}-$\mu$ and \poly{}-$\mu$) will have both multi-task and adaptation parameters equal to that of a single shared adapter, since after multi-task pretraining one can average the modules. 

%%%%%% 
% Preliminary Results for T0-11B
\iffalse
\subsubsection{Larger Scale Results}

Here we present preliminary results using a larger backbone, T0-11B \cite{sanh2022multitask}. We use the best configuration obtained on T0-3B for the 4 baselines in the table. Due to the added computational cost of training larger models, we only train for 25K (not 100K) steps. 

\begin{table}[h]
\small
\centering
\begin{tabular}{lccc}
\toprule
\textbf{Model} &  Multi-Task Params & Adaptation Params & Avg. Test \\
\midrule
\multicolumn{4}{l}{\textbf{T0 Dataset}} \\
\midrule
\multicolumn{4}{l}{\emph{Backbone T0-11B}} \\
\texttt{(IA)}$^3$ \cite{tfew} & 1.1M & 1.1M & 71.5$_{1.5}$ \\
\texttt{LoRA}  & 4.4M & 4.4M & 72.2$_{0.8}$ \\
% \poly{}-$z$ & 17M & 3.5K & 66.4$_{0.3}$ \\
\poly{} & 35.2M & 4.4M & 73.1$_{1.1}$ \\
% \rowcolor{lightgray}\polys{}-$z$ \emph{(64 h)} & 17M & 220K & 68.3$_{0.8}$ \\
\rowcolor{lightgray}\polys{} \emph{(8 h)} & 35.2M & 4.4M & \underline{73.2}$_{1.2}$ \\
\bottomrule
\end{tabular}
\vspace{10pt}
% \hspace{0.01\textwidth}
\caption{T0 dataset \cite{sanh2022multitask}, using T0-11B parameter backbone. We use the same evaluation procedure as before.}
\label{tab:11B}
\end{table}
\fi
% Preliminary Results for T0-11B
%%%%%% 

\subsection{Navigating the parameter efficiency / performance trade-off of tuning only the routing}

Here we provide additional results on how different routing based methods can be more expressive when only learning a new routing function (over \textit{frozen} modules) to adapt to a new task. 

\begin{figure}[h]
\centering
\includegraphics[width=0.9\linewidth]{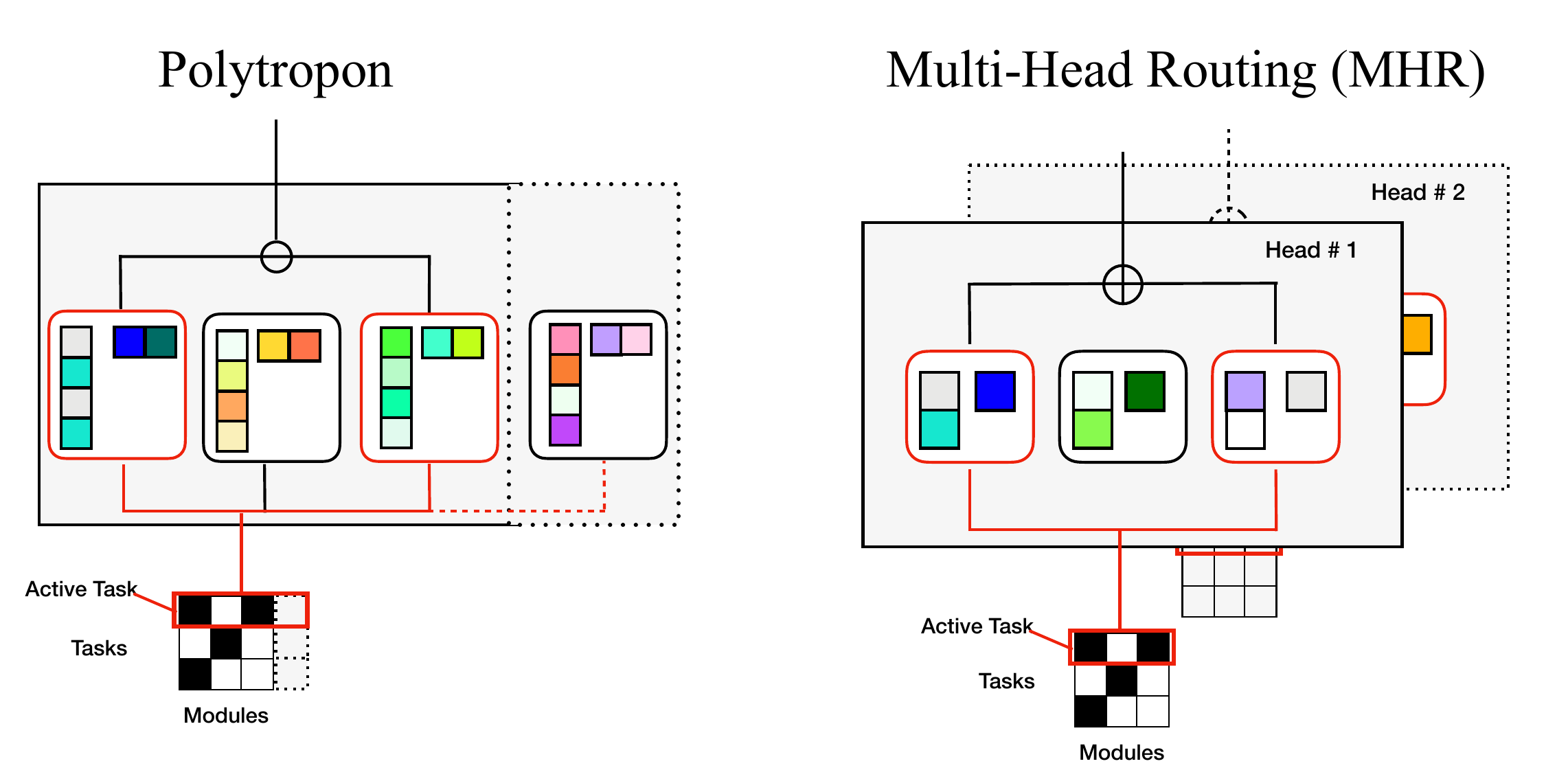}
\caption{Different ways to control the expressivity of routing based methods. \textit{Left : } In Polytropon, one can only add additional modules, resulting in a linear parameter increase. \textit{Right : } In \polys{}, additional heads only introduce routing matrices $\mZ$, resulting in a negligible parameter increase. %\textit{Right : } We also explore tying the module allocation $Z$ weights across adapters in the same layer for more parameter efficient adaptation.
} 
\label{fig:poly_vs_polys}
\end{figure}

In Fig. \ref{fig:poly_vs_polys} (left), we see that in order to build more expressive routing functions $\mZ$, in \poly{} one can only do so by increasing the number of skills at each layer. However, this has a significant impact on the number of multi-task parameters which much be kept in order to perform few-shot transfer. \polys{} on the other hand, can increase routing capacity in a much more parameter efficient way.

\subsubsection{On the granularity of routing tensor in \polys{}}
\label{sec:poly_granularity_appendix}
Here we provide additional results when modifying the granularity of $\tZ$ for \polys{}. We see that one can easily trade-off more parameters for better performance. 

\begin{figure}[h]
\centering
\includegraphics[width=0.59\linewidth]{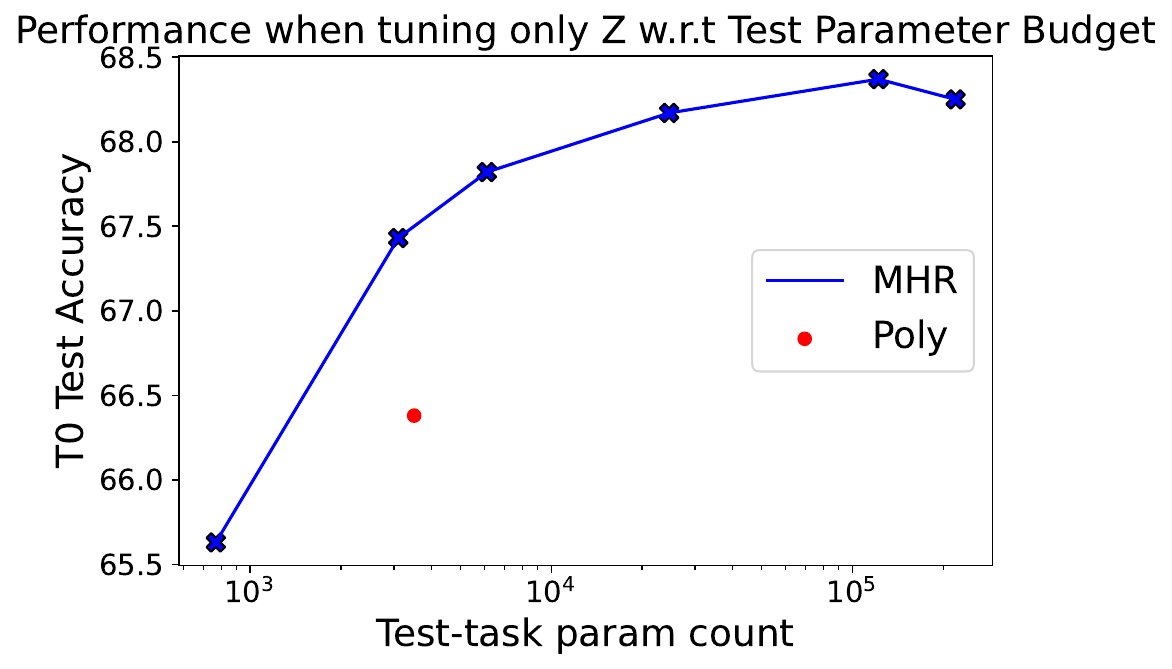}
\caption{Routing-Only Fine-Tuning (\polys{}-$z$)} 
\label{fig:poly_granularity_appendix}
\end{figure}

\newpage
\section{Broader Impact} 
% \TODO{Do we need this ? }

In our work, we focus on advancing parameter-efficient fine-tuning methods for cross-task generalization. While our research primarily addresses technical challenges and performance improvements, when applying such methods, it is crucial to consider the potential negative societal impacts. Specifically, we believe that prior to applying our proposed adaptation method, critically examining the potential biases and ethical implications of the underlying large language model, and the data itself must be properly addressed. This includes issues related to fairness, privacy, and the spread of misinformation. 

\end{document}